\documentclass[letterpaper, conference]{IEEEtran}
\IEEEoverridecommandlockouts
\usepackage{cite}
\usepackage{amsmath,amssymb,amsfonts}
\usepackage{algorithmic}
\usepackage{graphicx}
\usepackage{textcomp}
\usepackage{xcolor}
\usepackage{cleveref}
\crefname{figure}{Fig.}{Figs.}
\Crefname{figure}{Fig.}{Figs.}
\crefname{table}{Table}{Tables}
\Crefname{table}{Table}{Tables}
\crefname{equation}{Eq.}{Eqs.}
\Crefname{equation}{Eq.}{Eqs.}
\crefname{section}{Section}{Sections}
\Crefname{section}{Section}{Sections}
\usepackage{booktabs}
\usepackage{subcaption}
\usepackage{multirow}
\usepackage{microtype}

\newcommand{\sotatable}{
\begin{table*}[tb]
\centering
\caption{Comparison of results with state-of-the-art methods. Reported values represent average AUC-PRC over five runs, except for the teacher model, where both top-1 and average results are shown (with the latter used for comparisons). \textbf{Bold} values indicate the best AUC-PRC among student models for each dataset.}
\label{tab:sota}
\renewcommand{\arraystretch}{0.95} % Default value: 1
\begin{tabular}{@{}lcccccccccccc@{}}
\toprule
Dataset & \begin{tabular}[c]{@{}c@{}}Teacher\\ (max)\end{tabular} & \begin{tabular}[c]{@{}c@{}}Teacher\\ (avg.)\end{tabular} & Base & \begin{tabular}[c]{@{}c@{}}Base-\\ KD\end{tabular} & DKD & Att. & \begin{tabular}[c]{@{}c@{}}RKD-\\ D\end{tabular} & \begin{tabular}[c]{@{}c@{}}RKD-\\ A\end{tabular} & \begin{tabular}[c]{@{}c@{}}RKD-\\ DA\end{tabular} & Fitnet & DT2W & Ours \\ \midrule
UWaveGestureLib.All & 94.75 & 91.72 & 77.15 & 79.07 & 80.58 & 79.31 & 78.34 & 78.99 & 77.51 & 81.96 & 81.35 & \textbf{85.88} \\
Strawberry & 94.25 & 87.91 & 70.68 & 73.06 & 71.13 & 89.09 & 90.05 & 91.53 & \textbf{92.08} & 87.1 & 59.54 & 90.48 \\
Adiac & 70.61 & 65.79 & 37.5 & 45.75 & 46.36 & \textbf{49.21} & 41.93 & 43.3 & 39.94 & 37.67 & 43.66 & 45.07 \\
ItalyPowerDemand & 99.24 & 98.21 & 94.22 & 98.63 & 97.68 & 97.55 & 98.37 & \textbf{99.07} & 98.78 & 97.23 & 97.94 & 98.66 \\
yoga & 75.88 & 73.04 & 57.97 & 67.41 & 66.06 & 70.61 & 68.36 & 67.78 & 69.33 & 60.1 & 52.12 & \textbf{74.01} \\
Trace & 73.84 & 54.39 & 64.76 & 69.25 & 68.21 & 61.61 & 72.71 & 72.94 & 72.48 & 74.11 & 53.14 & \textbf{78.76} \\
ShapesAll & 72.34 & 71.04 & 29.82 & 35.34 & \textbf{44.46} & 35.11 & 30.27 & 30.98 & 31.98 & 35.97 & 33.55 & 40.69 \\
SwedishLeaf & 92.13 & 89.65 & 70.75 & 71.47 & 72.95 & 76.39 & 70.87 & 72.94 & 71.11 & 71.72 & 73.14 & \textbf{77.37} \\
FaceAll & 87.16 & 82.77 & 46.22 & 54.38 & 59.82 & 58.21 & 52.68 & 54.6 & 57.88 & 56.08 & 55.92 & \textbf{62.26} \\
MoteStrain & 83.89 & 80.75 & 82.74 & 81.87 & 81.39 & 82.58 & 83.02 & 83.53 & 82.76 & 78.16 & \textbf{84.21} & 84.13 \\
NonInvasiveFatalECG1 & 86.11 & 82.29 & 58.08 & 57.92 & 56.96 & 64.21 & 61.02 & 58.4 & 55.09 & 65.4 & 55.01 & \textbf{69.18} \\
NonInvasiveFatalECG2 & 91.71 & 86.7 & 71.2 & 74.5 & 66.69 & 73.26 & 73.25 & 73.51 & 73.41 & 71.13 & 72.64 & \textbf{77.47} \\ \hline
Avg. AUC-PRC & 85.16 & 80.36 & 63.42 & 67.39 & 67.69 & 69.76 & 68.41 & 68.96 & 68.53 & 68.05 & 63.52 & \textbf{73.66} \\
Wins & - & - & 0 & 0 & 1 & 1 & 0 & 1 & 1 & 0 & 1 & \textbf{7} \\
Lose & - & - & 12 & 12 & 11 & 11 & 12 & 11 & 11 & 12 & 11 & \textbf{5} \\
Avg. rank & - & - & 8.83 & 5.67 & 5.75 & 4.58 & 6.08 & 4.58 & 5.42 & 5.92 & 6.42 & \textbf{1.75} \\
Wins(with teacher) & - & \textbf{7} & 0 & 0 & 0 & 0 & 0 & 1 & 1 & 0 & 1 & 2 \\
Lose(with techer) & - & \textbf{5} & 12 & 12 & 12 & 12 & 12 & 11 & 11 & 12 & 11 & 10 \\
Avg. rank(with teacher) & - & 3.42 & 9.67 & 6.42 & 6.58 & 5.33 & 6.75 & 5.25 & 6.08 & 6.83 & 7.33 & \textbf{2.33} \\ \bottomrule
\end{tabular}%
\vspace{-12pt}
\end{table*}
}

\newcommand{\BasevsOursCombined}{
\begin{table}[tb!]
\centering
\caption{Performance comparison of our method with baseline methods across 12 UCR datasets.}
\hfill
\begin{subtable}{0.26\textwidth}
\centering
\caption{Comparison with student-base and base-KD}
\begin{scriptsize}
\setlength{\tabcolsep}{2.4pt} % Default value: 6pt
\begin{tabular}{@{}lccc@{}}
\toprule
 & Base & Base-KD & Ours \\ \midrule
Avg. AUC-PRC & 63.42 & 67.39 & \textbf{73.66} \\
Avg. AUC-ROC & 87.33 & 89.73 & \textbf{92.68} \\
Avg. Acc. & 61.15 & 64.25 & \textbf{69.12} \\
Win & 0 & 1 & \textbf{11} \\
Tie & 0 & 0 & \textbf{0} \\
Lose & 12 & 11 & \textbf{1} \\
Avg. Rank & 2.83 & 2.08 & \textbf{1.08} \\ \bottomrule
\end{tabular}
\end{scriptsize}
\label{tab:base_vs_ours}
\end{subtable}
\hfill
\hfill
\hfill\hfill
\begin{subtable}{0.20\textwidth}
\centering
\caption{Comparison with Fitnets\\}
\vspace{-0.115in}
\begin{scriptsize}
\setlength{\tabcolsep}{2.4pt} % Default value: 6pt
\begin{tabular}{@{}lcc@{}} \\ \toprule
 & Fitnet & Ours \\ \midrule
Avg. AUC-PRC & 68.05 & \textbf{73.66} \\
Avg. AUC-ROC & 90.09 & \textbf{92.68} \\
Avg. Acc. & 65.24 & \textbf{69.12} \\
Win & 0 & \textbf{12} \\
Tie & 0 & \textbf{0} \\
Lose & 12 & \textbf{0} \\
Avg. Rank & 2 & \textbf{1} \\ \bottomrule
\end{tabular}
\end{scriptsize}
\label{tab:fits_vs_ours}
\end{subtable}
\hfill
\label{tab:base_vs_ours_combo}
\vspace{-0.200in}
\end{table}
}

\newcommand{\overview}{
\begin{figure}[htbp]
\centerline{\includegraphics[width=0.49\textwidth]{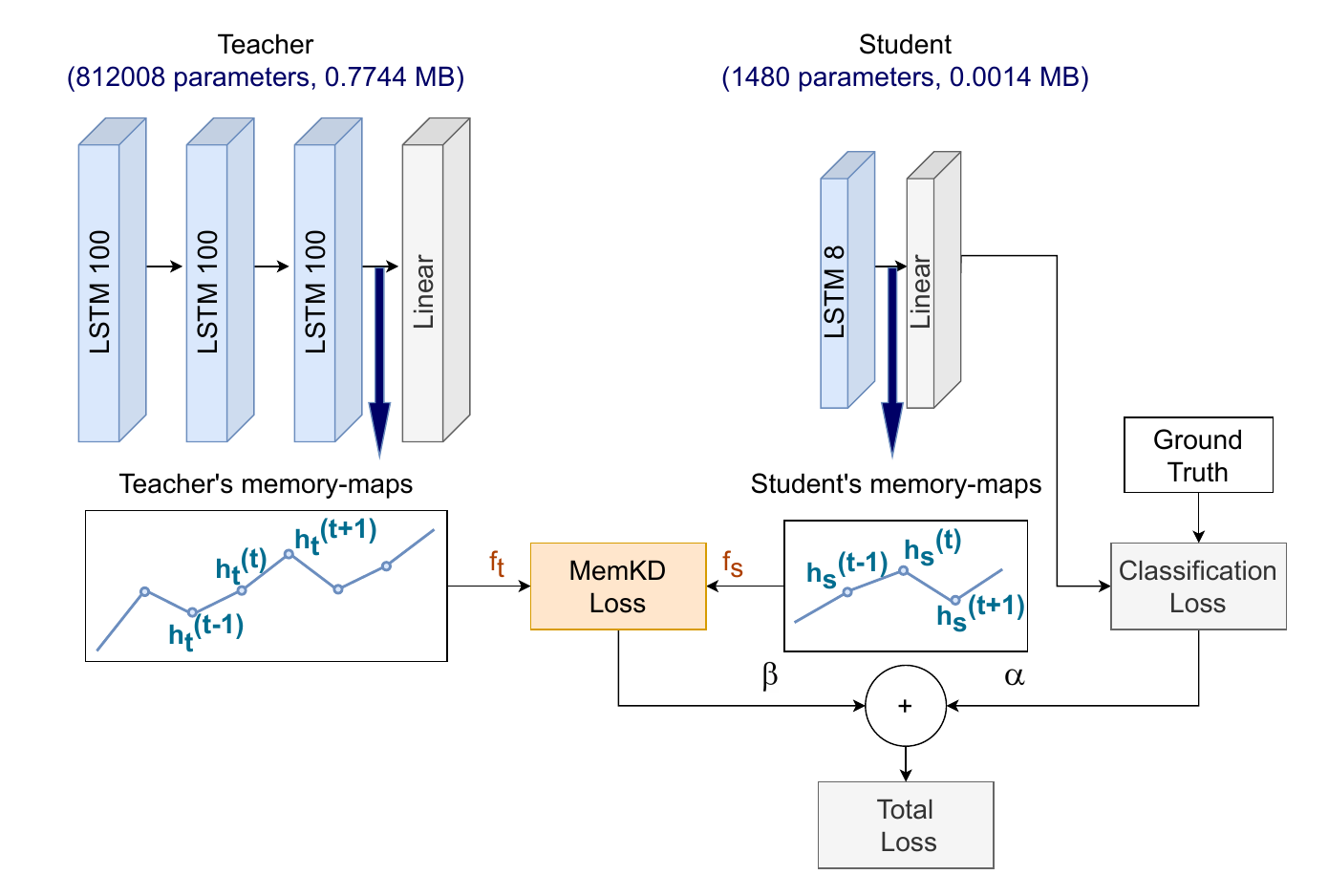}}
\looseness=-1
  \caption{An overview of the proposed knowledge distillation framework. The student learns target task by minimizing the classification loss while mimicking the memory retention of the teacher network using intermediate memory-maps.}
  \label{pic:overview}
    \vspace{-0.2in}
\end{figure}
}

\def\BibTeX{{\rm B\kern-.05em{\sc i\kern-.025em b}\kern-.08em
    T\kern-.1667em\lower.7ex\hbox{E}\kern-.125emX}}
\begin{document}

\title{MemKD: Memory-Discrepancy Knowledge Distillation for Efficient Time Series Classification
}

% \author{
% \IEEEauthorblockN{Nilushika Udayangani, Nandakishor Desai, and Marimuthu Palaniswami
% }
% \IEEEauthorblockA{Department of Electrical and Electronic Engineering, The University of Melbourne, Parkville, Victoria - 3010, Australia\\
% Email: hewadehigaha@student.unimelb.edu.au, nnandakisho@unimelb.edu.au, palani@unimelb.edu.au
% } 
% }

\author{\IEEEauthorblockN{Nilushika Udayangani, Kishor Nandakishor, and Marimuththu Palaniswami}
\IEEEauthorblockA{\textit{Department of Electrical and Electronic Engineering, University of Melbourne}  Parkville VIC 3052, Australia\\
hewadehigaha@student.unimelb.edu.au, nandakishor.desai@unimelb.edu.au, and palani@unimelb.edu.au}
% \and
% \IEEEauthorblockN{Kishor Nandakishor}
% \IEEEauthorblockA{\textit{Department of Electrical and} \\
% \textit{Electronic Engineering} \\
% \textit{University of Melbourne}\\
% Parkville VIC 3052, Australia \\
% nandakishor.desai@unimelb.edu.au}
% \and
% \IEEEauthorblockN{Marimuththu Palaniswami}
% \IEEEauthorblockA{\textit{Department of Electrical and} \\
% \textit{Electronic Engineering} \\
% \textit{University of Melbourne}\\
% Parkville VIC 3052, Australia \\
% palani@unimelb.edu.au}
% }
}
% \author{\IEEEauthorblockN{Nilushika Udayangani}
% \IEEEauthorblockA{\textit{Department of Electrical and} \\
% \textit{Electronic Engineering} \\
% \textit{University of Melbourne}\\
% Parkville VIC 3052, Australia \\
% hewadehigaha@student.unimelb.edu.au}
% \and
% \IEEEauthorblockN{Kishor Nandakishor}
% \IEEEauthorblockA{\textit{Department of Electrical and} \\
% \textit{Electronic Engineering} \\
% \textit{University of Melbourne}\\
% Parkville VIC 3052, Australia \\
% nandakishor.desai@unimelb.edu.au}
% \and
% \IEEEauthorblockN{Marimuththu Palaniswami}
% \IEEEauthorblockA{\textit{Department of Electrical and} \\
% \textit{Electronic Engineering} \\
% \textit{University of Melbourne}\\
% Parkville VIC 3052, Australia \\
% palani@unimelb.edu.au}
% }

\maketitle

\begin{abstract}
Deep learning models, particularly recurrent neural networks and their variants, such as long short-term memory, have significantly advanced time series data analysis. These models capture complex, sequential patterns in time series, enabling real-time assessments. However, their high computational complexity and large model sizes pose challenges for deployment in resource-constrained environments, such as wearable devices and edge computing platforms. Knowledge Distillation (KD) offers a solution by transferring knowledge from a large, complex model (teacher) to a smaller, more efficient model (student), thereby retaining high performance while reducing computational demands. Current KD methods, originally designed for computer vision tasks, neglect the unique temporal dependencies and memory retention characteristics of time series models. To this end, we propose a novel KD framework termed \textbf{Mem}ory-Discrepancy \textbf{K}nowledge \textbf{D}istillation~(\textbf{MemKD}). MemKD leverages a specialized loss function to capture memory retention discrepancies between the teacher and student models across subsequences within time series data, ensuring that the student model effectively mimics the teacher model's behaviour. This approach facilitates the development of compact, high-performing recurrent neural networks suitable for real-time, time series analysis tasks. Our extensive experiments demonstrate that MemKD significantly outperforms state-of-the-art KD methods. It reduces parameter size and memory usage by approximately 500 times while maintaining comparable performance to the teacher model. 
\end{abstract}


\section{Introduction}
\looseness=-1
Time series deep learning models are increasingly crucial across various domains, from financial forecasting to industrial process control. These models aim to strike a balance between high performance in handling complex sequential data and the need for efficient, lightweight solutions deployable on edge devices. Recurrent neural networks (RNNs), particularly LSTMs\cite{hochreiter1997long} and GRUs~\cite{cho2014learning}, excel at time series analysis due to their ability to model sequential data using hidden states. However, RNNs face challenges in real-world applications that require continuous data processing and real-time predictions. These limitations stem from the need to update numerous hidden states using nonlinear functions with millions of parameters, making it difficult to meet the low power and space requirements of edge devices such as smartphones and wearables.

Knowledge distillation (KD) transfers knowledge from a large teacher model to a smaller student model, maintaining performance while improving efficiency. Introduced in~\cite{ba_deep_2014} and extended by~\cite{hinton_distilling_2014} and~\cite{romero_fitnets_2014} for image classification, KD showed that smaller models could match or outperform deeper ones. Despite its broad applicability, KD remains understudied in time series analysis. Ay et al.~\cite{ay_study_2022} conducted the first study on KD for time series classification~(TSC) using fully convolutional networks, following the approach in~\cite{hinton_distilling_2014} to align output probability distributions. Subsequently, Ouyang et al.~\cite{ouyang_knowledge_2023} introduced a KD framework for TSC, extracting knowledge in both time and frequency domains by aligning output distributions in separate domains. However, these studies primarily adapt KD frameworks originally developed for computer vision tasks, without tailoring their loss functions to exploit the unique characteristics of time series data. Time series values, unlike visual data, exhibit unidirectional correlation, and during classification, short contiguous subsequences often contain much of the discriminative information~\cite{parvatharaju_learning_2021,ye_time_2009}. Consequently, direct adaptations of image-based approaches to time series may inadvertently disrupt the knowledge of these discriminative subsequences and their correlations. There are limited existing studies that explore KD specifically for time series, particularly using RNN teachers. These include matching the output probability distributions between the teacher and student~\cite{ma_knowledge_2022} and matching respective feature maps using adversarial learning~\cite{xu_kdnet-rul_2022,xu_contrastive_2022}. However, these methods primarily focus on different architectural knowledge transfer techniques aimed at transferring the superior performance of RNNs to other less complex model families, such as convolutional neural networks~(CNNs). 

% The critical role of memory retention in RNNs prompts an unexplored avenue in knowledge distillation: leveraging a teacher network's memory mechanisms to enhance knowledge transfer. This is a problem that none of the aforementioned studies have investigated. We explore the effectiveness of transferring memory process information between teacher and student recurrent neural networks, aiming to enhance student model performance and advance knowledge distillation techniques for sequential data analysis. To this end, we propose a novel KD framework, termed \textit{Memory-Discrepancy Knowledge Distillation~(MemKD)}, which transfers knowledge from RNNs based on their memory maintenance over input sequences. At the core of MemKD, we design a novel loss function that compels the student network to mimic expected changes in the teacher's memory retention over short contiguous subsequences. Our approach successfully distills knowledge from a large LSTM-RNN model to a small LSTM-RNN model, achieving a remarkable reduction in parameter size and memory usage by approximately 500 times while maintaining comparable performance to the teacher.
\looseness=-1
The critical role of memory retention in RNNs prompts an unexplored avenue in knowledge distillation: leveraging a teacher network's memory mechanisms to enhance knowledge transfer. This is a problem that none of the aforementioned studies have investigated, which brings us to the main topic of this paper: can a teacher network improve the performance of a student network by providing it with information about what it remembers? To this end, we propose a novel KD framework, termed \textit{Memory-Discrepancy Knowledge Distillation~(MemKD)}, which transfers knowledge from RNNs based on their memory maintenance over input sequences, as shown in~\cref{pic:overview}. At the core of MemKD, we design a novel loss function that compels the student network to mimic expected changes in the teacher's memory retention over short contiguous subsequences. Our approach successfully distills knowledge from a large LSTM-RNN model to a small LSTM-RNN model, achieving a remarkable reduction in parameter size and memory usage by approximately 500 times while maintaining comparable performance to the teacher. The main contributions of our work are as follows:
\begin{itemize}
\item MemKD: A novel knowledge distillation framework that efficiently compresses complex LSTM models for time series classification while preserving performance.
\item A novel loss function that captures and transfers RNN memory retention patterns across diverse shapes of subsequences in time series data.
\item A feature-level knowledge transfer method that enables seamless knowledge flow between RNNs with varying dimensional representations, overcoming traditional size-matching constraints.
\item Comprehensive empirical validation across multiple datasets, demonstrating MemKD's significant performance improvements over state-of-the-art KD methods in time series tasks.
\end{itemize}
The rest of the paper is organized as follows:~\Cref{section_related_works} reviews the relevant literature,~\cref{method} details the MemKD framework,~\cref{experiments} outlines the experimental methodology,~\cref{discussion} presents the results, and~\cref{conculsion} discusses the findings and future research directions.

\section{Related Works}\label{section_related_works}
KD fundamentally aims to extract rich knowledge from a teacher network and effectively transfer it to guide a student network's training.
To optimize knowledge quality, various methodologies have been proposed. Among these, logit-level KD techniques, which extract knowledge from softened output distributions~\cite{hinton_distilling_2014}, remain the most widely adopted. Advancing logit-level techniques, DKD~\cite{zhao_decoupled_2022} improves performance by decoupling logit-level knowledge from target and non-target class distributions. Beyond the final layer logits, Fitnets~\cite{romero_fitnets_2014} proposed the direct matching of features from intermediate layers. Transferring feature-level knowledge using attention maps has also been studied~\cite{zagoruyko_paying_2017,huang_like_2017}. Approaches such as VID~\cite{ahn_variational_2019} aim to maximize mutual information between the student and teacher networks by variational information maximization, while RKD~\cite{park_relational_2019} focuses on extracting relationships among input data samples. These methods are considered state-of-the-art in KD; however, they were originally developed and tested for computer vision tasks. As a result, their loss functions are tailored to capture visual features and may lack direct analogies for time series data. Recent works have only begun addressing this gap by adapting these image approaches for time series. For example, the work in ~\cite{ay_study_2022,ouyang_knowledge_2023} studied logit-level KD for time series, and DT2W~\cite{qiao_class-incremental_2023} extends Fitnets~\cite{romero_fitnets_2014} by replacing Euclidean distance with soft-dynamic time warping distance. Xu et al.~\cite{xu_kdnet-rul_2022,xu_contrastive_2022} utilized adversarial learning to compress RNN models into CNNs in their studies focusing on disparate architecture KD methods. 
However, possibility of knowledge extraction from RNN memory states has received limited research attention. The only closely related study is presented for time series forecasting in~\cite{murata_recurrent_2023}, which aligns the last hidden state of the teacher and the student directly. However, this approach falls short in two critical aspects: firstly, it primarily focuses only on enhancing performance without addressing model compression, and secondly, it imposes the limitation of requiring same architectures for both teacher and student networks. Preferred over direct matching of hidden states, our proposed approach extracts rich memory-state knowledge, enabling efficient transfer from large to compact RNN models without size-matching constraints, preserving critical temporal information while significantly reducing model size.

\section{Memory-Discrepancy Knowledge Distillation}\label{method}
\overview
\subsection{Problem Formulation}
A time series can be represented as \begin{math}
\boldsymbol{X} = \left\{\boldsymbol{x}^{(1)}, \boldsymbol{x}^{(2)},\cdots, \boldsymbol{x}^{(T)}\right\}
\end{math}, where $\boldsymbol{x}^{(i)} \in \mathbb{R}^n$, $i=1,\cdots,T$ and $\boldsymbol{X}\in \mathbb{R}^{T\times n}$.
A time series dataset $\mathcal{D}$, which includes M samples, is a collection of pairs $(\boldsymbol{X}_i, \boldsymbol{Y}_i), i= 1, \cdots,M$, where $\boldsymbol{Y}_i \in \mathbb{R}^C$ is the class label for $\boldsymbol{X}_i$. The TSC problem involves training a classifier $\mathcal{F}$, mapping $\boldsymbol{X}_i$ to $\boldsymbol{Y}_i$. KD aims to train a shallow student model as $\mathcal{F}$ using the knowledge from a pre-trained deep teacher model, minimizing the distillation loss:
\begin{equation}\label{eq_lkd}
L_{\mathrm{KD}} =\sum _{\boldsymbol{X}_i} \ell\left(f_t(\boldsymbol{X}_i),f_s(\boldsymbol{X}_i)\right)
\end{equation} where $f_t$ and $f_s$ represent functions for teacher and student models respectively and $\ell$ is a similarity measure used to match the metrics of the teacher and student models.

%\subsection{Memory-Discrepancy Knowledge Distillation}
\subsection{Formulating Knowledge from Hidden States}
Given an input time series sequence
\begin{math}
\boldsymbol{X}_i = \left\{\boldsymbol{x}^{(1)}, \boldsymbol{x}^{(2)}, \cdots, \boldsymbol{x}^{(T)}\right\}
\end{math}, the vanilla RNN processes each input time step $\boldsymbol{x}^{(t)} \in \mathbb{R}^n $ to maintain a time-varying memory state known as the hidden state $\boldsymbol{h}^{(t)} \in \mathbb{R}^m$. At time step $t$, the RNN updates hidden state $\boldsymbol{h}^{(t-1)}$ to $\boldsymbol{h}^{(t)}$ using both $\boldsymbol{x}^{(t)}$ and $\boldsymbol{h}^{(t-1)}$ as:

\begin{equation}\label{eq1}
\boldsymbol{h}^{(t)} = f(\boldsymbol{W}_h \boldsymbol{h}^{(t-1)} + \boldsymbol{W}_i \boldsymbol{x}^{(t)})
\end{equation}
where $\boldsymbol{W}_h$ and $\boldsymbol{W}_i$ are weight matrices and $f$ is a non-linear activation function which constrains the range of $\boldsymbol{h}^{(t)}$.
After processing the entire sequence at time step $T$, a linear layer can be trained upon last hidden state $\boldsymbol{h}^{(T)}$ to derive the output probability distribution for classification tasks. By unrolling past hidden states in~\cref{eq1}, each hidden state $\boldsymbol{h}^{(t)}$ at time step t can be expressed as a function of past input values up to time step t, \begin{math}
\left\{\boldsymbol{x}^{(1)}, \boldsymbol{x}^{(2)},\cdots, \boldsymbol{x}^{(t)}\right\}
\end{math} and the initial hidden state value $\boldsymbol{h}^{(0)}$.
The hidden state vector in an RNN represents the network's memory at each time step, encapsulating information about the input sequence processed up to that point. It evolves as the RNN processes the input sequence, capturing temporal dependencies and patterns in the data. This dynamic representation of memory and temporal information makes the hidden state a potentially valuable source of knowledge for KD frameworks. Let us consider the hidden state $\boldsymbol{h}^{(t+z)}$ at time step $t+z$, $z \in \mathbb{Z}$ . By unrolling each hidden state in~\cref{eq1} up to $\boldsymbol{h}^{(t)}$, it can be written as
\begin{math}
\boldsymbol{h}^{(t+z)} = f(\boldsymbol{h}^{(t)}, \boldsymbol{x}^{(t+1)}, \boldsymbol{x}^{(t+2)},\cdots, \boldsymbol{x}^{(t+z)})
\end{math} where $f$ is a nonlinear function.
Now consider the difference between any two hidden states $\boldsymbol{h}^{(t+z)}$ and $\boldsymbol{h}^{(t)}$ : 
\begin{equation}\label{eq_delatah}
\Delta \boldsymbol{h}^{t,z} = (\boldsymbol{h}^{(t+z)}-\boldsymbol{h}^{(t)})
\end{equation}
This can be regarded as response of memory to input values in subsequence 
\begin{math}
\left\{\boldsymbol{x}^{(t+1)}, \boldsymbol{x}^{(t+2)},\cdots,\boldsymbol{x}^{(t+z-1)}, \boldsymbol{x}^{(t+z)}\right\}
\end{math}, denoted $\boldsymbol{X}^{t:(t+z)}$ here. Although $\boldsymbol{h}^{(t+z)}$ is calculated by a non-linear transformation of $\boldsymbol{h}^{(t)}$ and values in $\boldsymbol{X}^{t:(t+z)}$, $\Delta \boldsymbol{h}^{t,z}$ can be regarded as deterministic to the input values in subsequence $\boldsymbol{X}^{t:(t+z)}$ when the previous history $\boldsymbol{h}^{(t)}$ is given. Thus $\Delta \boldsymbol{h}^{t,z}$ can reflect to what degree the model's hidden state (memory) is influenced by the input values in subsequence $\boldsymbol{X}^{t:(t+z)}$.

\subsection{Knowledge Distillation Objective}
We hypothesize that, after the occurrence of certain patterns in time series data, the model's tendency to memorize subsequent smaller occurrences or data patterns provides a valuable source of knowledge to extract. Based on this, we introduce a novel distillation loss function using the metric in~\cref{eq_delatah}, which quantifies an input subsequence's impact on model memory.
Our loss function aims to minimize the difference between these metrics among teacher and student. We define $f_t$ and $f_s$ in~\cref{eq_lkd} as:
% \begin{equation}\label{eq_ft}
% f_t = \left\{\frac{( \boldsymbol{h}_t^{(t+z)}-\boldsymbol{h}_t^{(t)})}{ \parallel \boldsymbol{h}_t^{(t)} \parallel}\right\}t,z\in [(T-1)], t+z\leq T
% \end{equation} 
% \begin{equation}\label{eq_fs}
% f_s = \left\{\frac{(\boldsymbol{h}_s^{(t+z)}-\boldsymbol{h}_s^{(t)})}{ \parallel \boldsymbol{h}_s^{(t)} \parallel}\right\}t,z\in [(T-1)], t+z\leq T
% \end{equation} 
\begin{equation}\label{eq_ft}
f_t = \left\{\frac{\Delta \boldsymbol{h}_t^{t,z}}{ \parallel \boldsymbol{h}_t^{(t)} \parallel}\right\}t,z\in [T-1], t+z\leq T
\end{equation} 
\begin{equation}\label{eq_fs}
f_s = \left\{\frac{\Delta \boldsymbol{h}_s^{t,z}}{ \parallel \boldsymbol{h}_s^{(t)} \parallel}\right\}t,z\in [T-1],t+z\leq T
\end{equation} 
where $\boldsymbol{h}_t$ and $\boldsymbol{h}_s$ represent hidden state vectors of teacher and student respectively. To ensure stability of the loss function and address scale differences between teacher and student hidden states, we normalize each difference by $\parallel \boldsymbol{h}^{(t)}\parallel$ in~\cref{eq_ft} and~\cref{eq_fs}. Further, to measure the difference between $f_t$ and $f_s$, we employ \textit{Smooth L1 loss} for our memory-discrepancy distillation loss:
\begin{equation}\label{eq_memkd}
L_{\mathrm{MemKD}} = \sum_{\boldsymbol{X}_i} \ell_{\mathrm{smoothL1}}\left(f_t(\boldsymbol{X}_i), f_s(\boldsymbol{X}_i)\right)
\end{equation}
where $L_{\mathrm{MemKD}}$ represents the proposed memory-discrepancy distillation loss. Smooth L1 loss combines quadratic and linear terms, offering advantages over both L1 and L2 losses: it's smoother than L1, less sensitive to outliers than L2, and helps prevent gradient explosion~\cite{girshick_2015_fast}. For shorter subsequences, we set $z=1$ and examine $\left\{\Delta \boldsymbol{h}^{t,1}\right\}_{t=1,\cdots,(T-1)}$, representing a measure of memory state changes due to each input value. This forms the memory-discrepancy distillation loss for shorter subsequences, $L_{\mathrm{MemKD-Short}}$. For longer subsequences, we use $z>1$ and randomly select subsets of $t$ and $z$. This allows $\left\{\Delta \boldsymbol{h}^{t,z}\right\}$ to capture impacts from diverse longer subsequences across the entire time series, forming $L_{\mathrm{MemKD-Long}}$. We examined the effects of these two terms separately and found that combining them yields the best performance.

% Our method calculates feature distillation loss indirectly, eliminating the need for additional alignment layers between teacher and student dimensions. This approach preserves semantic information while significantly reducing computational and space requirements. It offers a streamlined, efficient feature distillation process without compromising effectiveness or adding complexity during training.
Our method calculates the feature distillation loss indirectly, eliminating the need for additional alignment layers between the teacher and student dimensions~\cite{romero_fitnets_2014, huang_like_2017}. It is designed to avoid direct feature matching and to better preserve semantic information. This approach offers a streamlined and efficient feature distillation process without compromising effectiveness or adding complexity during training.

% \networkParamsReduced
\BasevsOursCombined
\sotatable
% \vspace{-0.1in}
\section{Experiments}\label{experiments}
% \vspace{-0.001in}
We used a three-layer LSTM (hidden size 100) as the teacher and a single-layer LSTM (hidden size 8) as the student, achieving a 500x compression (see~\cref{pic:overview}). For evaluation, we selected 12 datasets from the UCR-2015 archive~\cite{chen_ucr_2015}, categorized as short ($\le$150), medium (150-500), and long ($>$500) based on time series length~\cite{Xiao_2021_new, xing_2022_efficient}. The subset includes 4 datasets of each length category, with 7 having a large number of output classes~\cite{campos_2023_lightts}. We maintained original train/test splits, used 20\% of training data for validation, and preprocessed data by standardizing length to 100 and applying z-normalization.

We assessed model performance using area under the receiver operating characteristic curve~(AUC-ROC), average area under the precision-recall curve~(AUC-PRC), and accuracy, prioritizing AUC-PRC for its robustness to class imbalance. Performance comparisons utilized a win/tie/loss calculation and average-rank based on AUC-PRC. Teacher models were trained with five initializations per dataset, selecting the best based on validation AUC-PRC. Student models were trained using various KD methods: MemKD, Base-KD~\cite{hinton_distilling_2014}, FitNet~\cite{romero_fitnets_2014}, RKD~\cite{park_relational_2019}, Att~\cite{zagoruyko_paying_2017}, DKD~\cite{zhao_decoupled_2022}, and DT2W~\cite{qiao_class-incremental_2023}, plus a baseline 
(Base) without KD. For all student models involving KD, we trained them using a combination of the distillation loss and the classification loss (cross-entropy loss) as:
\begin{equation}
    L_{\mathrm{train}} = \alpha\times L_{\mathrm{CE}} + \beta\times L_{\mathrm{KD}}
    \label{total_loss}
\end{equation}
where $\alpha$ and $\beta$ decide the contribution of classification loss $L_{\mathrm{CE}}$ and distillation loss $L_{\mathrm{KD}}$ for the total train loss $L_{\mathrm{train}}$, respectively.
For all experiments, $\alpha$ is fixed at 1, while the optimal value of $\beta$ is selected via grid search.
Each student model is trained with five random initializations, and evaluation metrics are averaged across these runs. All models are implemented in PyTorch~\cite{paszke_2019_pytorch} using the same settings. Training is done with the Adam optimizer, a batch size of 32, and a maximum of 500 epochs, with early stopping based on validation loss. The initial learning rates were set to 0.01 for the teacher models and 0.1 for the student models.

\section{Discussion}\label{discussion}
% \noindent\textbf{Results and Discussion:}
% \subsection{Comparison with baseline methods}
We compared MemKD's performance with that of a student trained from scratch (Base) and vanilla knowledge distillation (Base-KD). The corresponding results across all 12 datasets are summarized in~\cref{tab:base_vs_ours}. It is evident that our method outperforms the Base model on all 12 datasets, establishing our proposed method as an efficient knowledge distillation framework. Additionally, the average performance metrics improved by approximately 5-10\% compared to the Base model. MemKD also outperforms vanilla knowledge distillation on approximately 90\% (11 wins) of the datasets.
Since MemKD is also a feature distillation method, we compared it with vanilla feature distillation (Fitnets), and the summarized results are presented in~\cref{tab:fits_vs_ours}. MemKD outperforms Fitnets on all datasets, achieving 2-5\% improvement in the average performance metrics compared to Fitnets. These results demonstrate that our loss function better exploits knowledge from the teacher’s feature layers, rather than providing direct hints from those layers, while also eliminating the need for additional regression layers. Table~\ref{tab:sota} presents the evaluation results of different KD variants across 12 datasets. We included win/tie/loss calculations both with and without the teacher model. The proposed method, MemKD, consistently outperforms all other distillation objectives in approximately 60\% of the datasets. The average rank of MemKD is 1.75, which is significantly lower than that of other methods, indicating that MemKD delivers competitive performance across the remaining datasets as well. None of the other methods were able to win at least 10\% of the datasets, nor did they achieve ranks below 4. Additionally, our method shows at least a 4\% improvement over other methods in terms of average AUC-PRC. All KD methods report lower ranks compared to the Base model, indicating that they all benefit from knowledge distillation. Furthermore, MemKD achieves the lowest average rank even in the calculations that include the teacher model. It is the only method to exceed the teacher in terms of average rank.

\section{Conclusion}\label{conculsion}
We introduced MemKD, a novel KD framework specifically tailored for deep time series recurrent networks. MemKD focuses on how memory evolves within RNNs and uses information about memory retention over short contiguous subsequences to improve student model performance. Using a unique loss function that captures memory retention discrepancies, MemKD effectively transfers knowledge from a large LSTM model to a much smaller one. Experiments on 12 datasets show that MemKD outperforms state-of-the-art KD methods, significantly reducing model size and memory usage while maintaining performance. Future work will focus on optimizing MemKD further and exploring its performance under various compression levels.

% We introduced Memory-Discrepancy Knowledge Distillation (MemKD), a novel KD framework specifically tailored for deep time series recurrent networks. We investigated how memory differs and evolves within RNNs and whether we can use information about memory retention over short contiguous subsequences to improve a student model. Using a unique loss function that captures memory retention discrepancies, MemKD effectively transfers knowledge from a large LSTM model to a much smaller one. Experiments on 12 datasets show that MemKD outperforms state-of-the-art KD methods, significantly reducing model size and memory usage while maintaining performance. Future work will aim to optimize MemKD further and explore performance under various compression levels.

\section*{Acknowledgment} \looseness=-1 This research was supported by The University of Melbourne’s Research Computing Services and the Petascale Campus Initiative.

% \clearpage
\bibliographystyle{IEEEtran}
\bibliography{name}

\vspace{12pt}

\end{document}